\documentclass{article}
\usepackage{amsmath,graphicx,mlspconf}
\usepackage{lipsum}
\usepackage{algorithmicx}
\usepackage{graphicx}
\usepackage{float}
\usepackage{color}
\usepackage{url}
\usepackage{algorithm}
\usepackage{algpseudocode}
\usepackage{booktabs} 
\usepackage{multirow}
\usepackage{siunitx}
\usepackage{comment}

\title{Unsupervised Domain Adaptation with Copula Models}
\name{Cuong D. Tran$^1$, Ognjen (Oggi) Rudovic$^2$, Vladimir Pavlovic$^1$}
\address{$^1$Department of Computer Science, Rutgers University, Piscataway, NJ, USA\\
            $^2$MIT Media Lab, Cambridge, MA, USA           \\
            {\tt\small \{cuong.tran,vladimir\}@cs.rutgers.edu, orudovic@mit.edu}}
\begin{document}
\newcommand*{\Scale}[2][4]{\scalebox{#1}{$#2$}}%
\newcommand*{\Resize}[2]{\resizebox{#1}{!}{$#2$}}%
	%
	
	\maketitle
	\begin{abstract}
		We study the task of unsupervised domain adaptation, where no labeled data from the target domain is provided during training time. To deal with the potential discrepancy between the source and target distributions, both in features and labels, we exploit a copula-based regression framework.  The benefits of this approach are two-fold: (a) it allows us to model a broader range of conditional predictive densities beyond the common exponential family; (b) we show how to leverage Sklar's theorem, the essence of the copula formulation relating the joint density to the copula dependency functions, to find effective feature mappings that mitigate the domain mismatch. By transforming the data to a copula domain, we show on a number of benchmark datasets (including human emotion estimation), and using different regression models for prediction, that we can achieve a more robust and accurate estimation of target labels, compared to recently proposed feature transformation (adaptation) methods.

		
		
	\end{abstract}
	\section{Introduction}
	\label{sec:intro}

	Domain adaptation is a key task in machine learning, in which one tries to learn a predictive model for a target domain by exploiting labeled data from different but related domains (often called the source domains). This approach is of critical importance when the number of labeled data in the target domain is limited and the process of labeling, necessary to create target-specific models, is expensive and slow. A simple solution of training a target model using only the source domains (e.g., applying the source model on the target) is, however, bound not to perform well due to the differences in the data distributions between the two domains (e.g., two subjects in emotion modeling tasks). Thus, by jointly learning from the scarce target and abundant source data, we can design an adaptation scheme from the source to target data, resulting in target-specific predictive models. These are expected to perform better when applied to the task-specific target domains, compared to generic (not adapted) models. 
	
	Existing domain adaptation works focus on two main strategies: an instance-based approach or a feature-based strategy. Instance-based strategies leverage a common assumption of covariate shift, i.e., that the underlying conditional density of outputs given covariates remains the same between two domains.  Based on that assumption, to correct the bias of the training source samples one re-weights each instance by a ratio of the target input density over the  source input density \cite{DBLP:conf/nips/KanamoriHS08,Shimodaira2000227,DBLP:conf/icassp/YamadaSM09}.  On the other hand, feature-based domain adaptation methods, c.f., \cite{Daume:2010:FES:1870526.1870534}, aim to construct domain-specific mappings (features) of the inputs across the domains that are most relevant for prediction in the target domain, implicitly assigning different contribution to samples from the two domains. While these approaches have proven to be effective, they require that samples in both source and the target domain be labeled.
	
	To avoid the requirement for labeled target data, unsupervised domain adaptation with feature-transformation is employed. The premise of the approaches in this context is that to reduce the dis-similarity between the domains one must learn the transformed input features invariant across domains, making the samples from different domains similar (i.e., aligned) to each other in density but still allowing the transformed samples to accurately predict labels on the labeled source domain \cite{DBLP:conf/iccv/BaktashmotlaghHLS13,DBLP:journals/tnn/PanTKY11}.  A simple measure of dis-similarity between the input samples from different domains uses the squared loss between covariance matrices from two domains, as in the CORAL \cite{DBLP:journals/corr/SunFS15} approach. Surprisingly, this simple method outperformed many prior competitive approaches on challenging object recognition tasks. However, for the adaptation, CORAL considers only second-order statistics, which could be difficult to estimate reliably from high-dimensional input data - which we address using the copula functions.



A common challenge faced by domain adaptation approaches is how to efficiently represent, and then align, the multivariate distributions across different domains. This task depends on the joint density of the inputs $x_1,\ldots,x_D$ and output $y$, $p(x_1,\ldots,x_D,y)$, which simultaneously and indistinguishably encode the multivariate dependency among factors and the priors/marginals in each factor, $p(x_i), i=1,\ldots,D$ and $p(y)$.  Instead of considering this typical joint representation, it may be more beneficial to consider a representation of joint dependencies that separates the notion of dependency among variables from their marginal distributions, which may differ across domains. The notion of Copula functions \cite{DBLP:journals/kybernetika/Sklar73} is a powerful statistical tool that allows us to decompose such distributions into a product of marginal distribution for each dimension and a copula function, $p(x_1,\ldots,x_d,y) = p(y) \prod_{i=1}^D p_{i}(x_i) c(F_1(x_1) ,\ldots,F_D(x_D),F_y(y))$, where $F_i(\cdot)$ denote the marginal cumulative distribution functions (CDFs). The copula function $c(\cdot,\ldots,\cdot)$ encodes the essential, marginal-free, dependence among variables. For instance, using this idea, \cite{DBLP:journals/corr/abs-1301-0142} proposed to augment the target model by exploiting the similarity between each component of target and source domain, including the marginals and the copula. However, this approach still requires some labeled target data during the training process. In case when no labeled target data are given, an alternative approach is to align the copula densities~\cite{DBLP:dblp_conf/icassp/BayestehtashkSB16}, thus, entirely discarding the source labels. However, that feature transformation is bound to perform inadequately on many practical problems, as will be demonstrated here, because it fails to make use of the valuable source data labels. 
    
We propose a novel domain adaptation framework using the copula representation that can alleviate the aforementioned issues. In particular, our proposed domain adaptation approach, 
based on copulas for the regression task, has the following benefits:
	\begin{itemize}
		\item The copula-based regression can represent more complicated conditional density relationships than a simple multivariate Gaussian, often assumed by existing approaches. 
		\item The copula-based feature transformation can match input data from two domains more closely than the frequently used affine transformation on raw input features, without the need for labeled target data. 
	\end{itemize}
We show that this proposed approach outperforms other unsupervised domain adaptation techniques on several benchmark datasets. We do so on a challenging task of predicting emotional responses of different subjects as a prototypical domain adaptation paradigm. The domains in that context represent different subjects and the task is to adapt the models trained on a subset of subjects, whose emotional responses have been labeled by expert annotators, to other subjects where no labeled data is available.  The subject specificity of emotional responses is the key challenge, which we address using the proposed unsupervised domain adaptation approach.

The paper is organized as follows. In Sec.~\ref{sec:copulas}, we first provide a brief introduction to copulas and a regression model based on copulas. Then, we introduce our domain adaptation model in Sec.~\ref{sec:model}. Experimental results that demonstrate the utility of the proposed approach on real datasets are reported in Sec.~\ref{sec:experiments}.

	\section{Problem formulation}\label{sec:copulas}
		\label{sec:format}
	The setting of unsupervised domain adaptation consists of data from two related but not identical domains: the source and target domain. We are interested in learning a regression model that can predict well on the target domain. During training, we are given a set of $n_S$ labeled source samples $\{\boldsymbol{X_S}, y_S\} \sim p_S(x,y)$ and a set of $n_T$ unlabeled target data $\boldsymbol{X_T} \sim p_T(x)$. The goal here is to use all labeled source and unlabeled target samples to build an adapted model $M_{S \to T}$ that predicts the true outcome $y^*_T$ on some target input data $\boldsymbol{X^*_T} $. In most cases, the dataset shift $p_T(x,y) \neq p_S(x,y)$ leads to the model trained on only the source data to underperform on the target domain. One of the main reasons for this is the gap between the input distributions $p_T(x) \neq p_S(x)$, i.e., the covariate shift. Thus, to obtain an accurate adapted model $M_{S \to T}$, one needs first to reduce that gap, while making sure that this does not affect adversely the model's predictive performance. In what follows, we briefly introduce the notion of copula functions and then demonstrate how the aforementioned task can be accomplished using this approach.

\subsection{Copulas}
	Given a set of $D$ input variables $\{x_i\}^D_{i=1}$ and an output variable $y$ with associated CDF: $F_{1},\ldots,F_{D}, F_{y}$, Sklar's theorem \cite{DBLP:journals/kybernetika/Sklar73} states that:
\begin{multline}  \exists  \ C: [0,1]^{D+1} \to [0,1]  \\
	\text{s.t} :  F(x_1,\ldots,x_D,y)=C(F_{1}(x_1) ,\ldots,F_{D}(x_D),F_{y}(y) )
\end{multline}
where $F(x_1,\ldots,x_D,y)$ is the joint CDF and the function $C$ is called a copula. Then, $C$ is unique iff $x_i,i=1,2,\ldots,D $ and $y$ are continuous random variables \cite{Nelsen:2006:IC:1204326}. Under this assumption, one can relate the joint density probability function (PDF) of $ y \  \text{and }x_i$ with the copula density $c$ as follows:
{\small \begin{align}  
	\lefteqn{p(x_1,\ldots,x_D,y)} \nonumber \\
    &= c(F_1(x_1) ,\ldots,F_D(x_D),F_y(y)) \cdot p(y) \prod_{i=1}^D p_{i}(x_i)  \label{maineq} \\
    &= c(u_1,\ldots,u_D,v) \cdot p(F^{-1}_y(v)) \prod_{i=1}^D p_{i}(F^{-1}_i(u_i)),
\end{align}}
where $u_i=F_i(x_i); z_i= \Phi^{-1}(u_i) ; \forall i$ and $v=F_y(y) ; w=\Phi^{-1}(v)$. Eq.~\ref{maineq} suggests that the joint PDF can be decomposed using two sets of factors:  independent marginals $p_{i}(x_i),p(y)$ and a copula density function $c(u_1,\ldots,u_D,v) = \frac{\partial^{D+1} C(u_1,\ldots,u_D,v)}{\partial u_1 \cdots \partial u_D \partial v}$ that models the essential dependency among variables, regardless of their marginal densities. 

Different forms of $C$ determine different copula models.  Typically, $C$ are parameterized using a small set of parameters, forming the family of Archimedean copulas, c.f.,~\cite{Nelsen:2006:IC:1204326}.  Another possibility is to assume that 
\begin{equation}\label{eq:gcop}
	c(u_1,\ldots,u_D,v |\boldsymbol{R}   )= | \boldsymbol{R} |^{-\frac{1}{2}} e^{-\frac{1}{2}  \boldsymbol{\tau}^T ( \boldsymbol{R}^{-1}-\boldsymbol{I}) \boldsymbol{\tau} } ,
\end{equation}
which defines the family of Gaussian copulas. Here $\boldsymbol{R}$ 
is the correlation matrix capturing the dependency among target variables, $\boldsymbol{I}$ is the unit matrix, and $\boldsymbol{\tau}=[z_1,\ldots,z_D,w]^T$. $\boldsymbol{R}$ can easily be estimated from data samples using maximum likelihood estimation \cite{choros2010copula}.

	\section{Domain adaptation with Gaussian copula models}\label{sec:model}
	
	Our goal is to learn the target and source feature mappings, $\phi_T(x) and \phi_S(x)$, respectively, that satisfy two conditions: (1) they should  minimize dissimilarity between the two domains, (2) the transformed (common) features $\phi_S(x)$ should retain the most discriminative information regarding $y$. Formally, this can be written as the following optimization problem: 
\begin{equation} \Scale[0.92]{ \min_{ \phi_T(.), \phi_S(.) } \mathcal{D} ( \phi_T( \boldsymbol{X_T} ); \phi_S(\boldsymbol{X_S} )  ) + \lambda \cdot g(y_S; \phi_S(\boldsymbol{X_S} ) )  }  \label{eq1}, \end{equation}	
where $\mathcal{D}$ is a distance function that measures the difference between probability measures how well the transformed input data explains the outputs. Since we do not have access to any labeled target data, we use $y_S; \phi_S(\boldsymbol{X_S} )$ to measure the correlation between output and input, where $\lambda >0 $ is a regularization parameter. Details of mapping functions $\phi_T(.), \phi_S(.) $, distribution distance function $\mathcal{D}$ and function $g$ are given in what follows.
	
\subsection{Mapping functions }
	
Let us consider the following domain specific mapping functions: target ($\boldsymbol{Z_T}=\Phi^{-1}(F_T( \boldsymbol{X_T} ))$), and source $(\boldsymbol{Z_S} =\Phi^{-1}(F_S( \boldsymbol{X_S} )))$. If we assume that $\boldsymbol{X_T}$ and $\boldsymbol{X_S}$  follow a Gaussian copula density, then  transformed features $ \boldsymbol{Z_T},\boldsymbol{Z_S}$ are multivariate Gaussians. Remember also that when the copulas of $x$ are Gaussian, the joint distribution over input $p(x)$ might not be Gaussian. Therefore, assuming a Gaussian copulas is not restricted to the input distributions. At this point, one might be interested in mapping two Gaussians by learning a common subspace so that the Gaussian samples can easily be matched (see Sec. \ref{stein}). To this end, we define the following feature mapping:

	$$\phi_T(x)=\boldsymbol{W^T} \Phi^{-1}(F_T(x)) $$
	
	$$\phi_S(x)= \boldsymbol{W^T} \Phi^{-1}(F_S(x)) $$
	Where $\boldsymbol{W} \in R^{D \times p}$ ($D \geq p$) is an orthogonal matrix, i.e $\boldsymbol{W^T} \boldsymbol{W}=\boldsymbol{I}$ that defines the common subspace between $\boldsymbol{Z_T}$ and $\boldsymbol{Z_S}$.
	
	\subsection{The distance function $\mathcal{D}$ } 
    \label{stein}
	Under Gaussian copula assumption on both domains, $ \boldsymbol{Z_T} , \boldsymbol{Z_S} $ are multivariate Gaussian with the same zero mean. Let their correlation matrices be $\boldsymbol{R_T}, \boldsymbol{R_S} $, respectively. As a result, $ \boldsymbol{W^T} \boldsymbol{Z_T} $ and   $ \boldsymbol{W^T} \boldsymbol{Z_S}$ are Gaussians with the same zero mean, and correlation matrices: $\boldsymbol{W^T} \boldsymbol{R_T} \boldsymbol{W}$ and $\boldsymbol{W^T} \boldsymbol{ R_S} \boldsymbol{ W} $, respectively. The problem turns out to be the second order matching between the two multivariate Gaussians that have the same mean.
One possible solution for Gaussian alignment is based on the KL divergence, as proposed in \cite{DBLP:dblp_conf/icassp/BayestehtashkSB16}. Yet, the KL divergence is asymmetric, and it is also sensitive to outliers because it involves the ratio of two densities, \cite{DBLP:journals/jmlr/Abou-MoustafaF12}. To avoid these issues, we use the Stein divergence \cite{DBLP:journals/corr/HerathHP16a}, $\mathcal{D_{S}}$, between the covariance matrices, $\boldsymbol{W^T} \boldsymbol{Z_T}$ and $\boldsymbol{W^T} \boldsymbol{Z_S}$, which is given by: 
	
	\vspace{-0.5cm}
	\begin{multline} 
\hspace{-1em} \mathcal{D_{S}} \bigg[ \mathcal{N}_T (\boldsymbol{W^T} z| 0; \boldsymbol{W^T} \boldsymbol{R_T}  \boldsymbol{W}), \mathcal{N}_S (\boldsymbol{W^T} z|0;\boldsymbol{W^T} \boldsymbol{R_S} \boldsymbol{W})    \bigg]   \   \     \  \      \   \     \  \   \\
	= \log \det \big( \frac{\boldsymbol{W^T} \boldsymbol{R_T}  \boldsymbol{W}  +\boldsymbol{W^T} \boldsymbol{R_S} \boldsymbol{W}}{2}\big) \\
    -\frac{1}{2} \log \det ( \boldsymbol{W^T} \boldsymbol{R_T} \boldsymbol{W}) (\boldsymbol{W^T} \boldsymbol{R_S}  \boldsymbol{W}).
    \end{multline}
Because we apply the Gaussian copula transformation to the input features, the second order statistics become the only necessary stats, and then we only need to consider some distance between Gaussian densities. 


\subsection{The alignment function $g$ } 
	


To make sure that, under the Copula transformation, the input features still can explain well the output, we require the mutual information (MI) between the output and transformed inputs to be maximized. The MI is used as a measure of dependency between random variables, and maximizing the MI between input and output variables is equivalent to minimizing the Bayes error \cite{fano1961transmission}. The MI between the output $y$ and input features $x$ is defined as

\begin{equation} MI(x,y)=\int \int p(x,y) \log \frac{p(x;y)}{p(x)p(y)} dxdy \end{equation}

However, MI can be sensitive to outliers \cite{Torkkola:2003:FEN:944919.944981}. Furthermore, since it involves the density ratio, it can go to infinity when either $p(x)$ or $p(y)$ is close to zero. Therefore, instead of maximizing the MI, we can maximize its lower bound, called the quadratic mutual information (QMI) \cite{Torkkola:2003:FEN:944919.944981}:
\begin{equation}
	QMI(x,y)=\int \int  \big( p(x,y)-p(x)p(y ) \big)^2 dx dy.
\end{equation}

The empirical estimation of $QMI(x,y)$ \cite{4928} using a set of drawn samples $\{x_i,y_i \}_{i=1}^n$ is given by:
\begin{equation}
	QMI(x,y)\approx \frac{1}{n^2} \text{trace} ( \boldsymbol{K_x} \boldsymbol{L} \boldsymbol{K_y}\boldsymbol{L} ),
\end{equation}
where $\boldsymbol{K_x}$ is the kernel matrix between input features, $\boldsymbol{K_y}$ is the kernel matrix between outputs, and $\boldsymbol{L}=\boldsymbol{I}-\frac{1}{n}e e^T $ is a constant centering matrix.  Note that since in our model no labeled target data is given during feature learning we can only maximize the QMI between transformed source input features and source outputs.


\subsection{Overall optimization problem}
	
Using the mapping function (Sec.3.1), the Stein distance (Sec.3.2), and the QMI between $y_S$ and $\boldsymbol{W^T} \boldsymbol{Z_S}$ (Sec.3.3), we obtain the following objective:
	\begin{multline} \min_{\boldsymbol{W}}  \ F(\boldsymbol{W})=\log \det \bigg( \frac{ \boldsymbol{W^T}  \boldsymbol{R_T} \boldsymbol{W}+ \boldsymbol{W^T}  \boldsymbol{R_S} \boldsymbol{W}}{2}\bigg)- \\
	\frac{1}{2}\log \det \big(\boldsymbol{W^T}  \boldsymbol{R_T} \boldsymbol{W} \boldsymbol{W^T} \boldsymbol{R_S} \boldsymbol{W} \big)
	-\lambda  \cdot \text{trace}(\boldsymbol{K_y}\boldsymbol{L} \boldsymbol{K}_{\boldsymbol{W^T}z} \boldsymbol{L})   \\   
	\text{s.t} \  \  \boldsymbol{W}^T \boldsymbol{W}=\boldsymbol{I} \label{main_obj}
	\end{multline}
The orthogonal constraint over $\boldsymbol{W}$ ensures that the projected data from two domains lie on a common subspace defined by $\boldsymbol{W}$. We employ the ManOpt toolbox~\cite{manopt} for the optimization problem above. After we obtain $\boldsymbol{W}_*$, we can train various regression models on the transformed features $\{\boldsymbol{W^T_*} \boldsymbol{Z_S}, y_S \}$. Then, the learned models will be used to predict the label of unseen transformed target samples, i.e $\{\boldsymbol{W^T_*}  \boldsymbol{Z^*_T}$ \}. The proposed domain adaptation framework is summarized in Alg.1.

\begin{algorithm}
		\caption{Unsupervised domain adaptation with Copula} 
        \begin{algorithmic}
        \Statex \textbf{Input:} $\boldsymbol{X_S}, y_S, \boldsymbol{X_T} $ and testing point $\boldsymbol{X^*_T} $ 
           \Statex	\textbf{Output:} Prediction $y^*_T$  of $\boldsymbol{X^*_T} $ 
           \vspace{0.1cm}
           \Statex Step1: Compute $\boldsymbol{ Z_T}=\Phi^{-1}(F_T(\boldsymbol{X_T)} ); \boldsymbol{Z_S} =\Phi^{-1}(F_S( \boldsymbol{X_S)})$ and $\boldsymbol{R_T}=\text{cov} (\boldsymbol{Z_T}); \boldsymbol{R_S}=\text{cov}(\boldsymbol{Z_S})$.
           
      \Statex Step 2:  Solve the optimization problem in Eq. \ref{main_obj} to obtain $\boldsymbol{W_*}.$
      
      \Statex Step 3: Compute the transformed features $\boldsymbol{W_*^T} \boldsymbol{Z_S} $ and $\boldsymbol{W_*^T}  \boldsymbol{Z^*_T} $.
      
      \Statex Step 4: Learn a regression model using $\{\boldsymbol{W_*^T} \boldsymbol{Z_S}, y_S \}$ and apply the models to predict $y^*_T$ of $\boldsymbol{W_*^T}  \boldsymbol{Z^*_T}. $
      
        \end{algorithmic} 
	\end{algorithm}

\section{Experiments}\label{sec:experiments}

\begin{table*}[tbh]
\footnotesize
	\begin{center}
    \caption{Comparison among models over the  UCI datasets. \textcolor{red}{Red} indicates the best result, followed by the second best in \textcolor{blue}{blue}. All models were evaluated based on NMSE values, the lower is the better. }
			\label{tb:uci}
				\hfill{}
				\begin{tabular}{lccccccccc} 
					\toprule
					& GPR & GCR & CT-GPR & CT-GCR  & NPRV & UNPRV & CORAL-GCR  & DA-GPR & DA-GCR  \\ 
                    \midrule
					\textbf{Wine} & 1.49 (0.58) & 1.81(0.27) & 0.93(0.03) & 0.90(0.04) & \textcolor{red}{\textbf{0.84(0.07) }} &  \textcolor{blue}{\textbf{0.85(0.07) }} &1.04(0.04) & 0.88(0.05) & 0.87(0.05) \\ 
                    
					\textbf{Airfoil Noise} &1.25(0.10) &0.70(0.02) & 0.70(0.08) & 0.65(0.04) & 1.12(0.03) & 1.12(0.03) & 0.96(0.09) & \textcolor{red}{\textbf{ 0.52 (0.08) }}  & 0.56 (0.03)  \\ 
                    
                    \textbf{Parkinsons} & 1.59  (0.02) & 1.41 (0.05) &  1.44(0.04)  &   \textcolor{blue}{ \textbf{ 1.18(0.04)}} & 1.46(0.11 ) & 1.21(0.03) & 3.52(0.29) & 1.47 (0.09) &  \textcolor{red}{ \textbf{ 1.13 (0.07)}} \\ 
                    \midrule
                       
                    \textbf{Av.Rank} & 8.3 & 5.8 & 5.2 &  \textcolor{blue}{\textbf{3.3}} & 4.8 & 4.2 & 7.3 & 4.0 & \textcolor{red}{\textbf{2.0}}  \\ 
					\bottomrule	
			\end{tabular}
			\hfill{}
	\end{center}
\end{table*}

In this section, we first study the behavior of our proposed domain adaptation methods over several publicly available UCI datasets \cite{Lichman:2013}. Subsequently, we highlight the benefits of the proposed method on the continuous emotion estimation tasks. For comparisons, we consider three baseline regression methods trained using only the source data: Gaussian Process Regression (GPR)\cite{Rasmussen:2005:GPM:1162254}, Gaussian Copula Regression (GCR) \cite{DBLP:conf/acl/WangH14}, and Non-parametric Regular Vines (NPRV) \cite{DBLP:journals/corr/abs-1301-0142}\footnote{In all kernel-based methods, we employ the radial basis function kernel.}. To compare the adaptation performance, we use two recently proposed feature transformation techniques: CORAL \cite{DBLP:journals/corr/SunFS15} and the Copula-based Transformation (CT) \cite{DBLP:dblp_conf/icassp/BayestehtashkSB16}. We evaluate these adaptation approaches by applying the above-mentioned regression models to the features adapted using the introduced transformations (CORAL, CT and the proposed copula-based DA). We also compare the performance of the NPRV model under the unsupervised domain adaptation setting (UNPRV) \cite{DBLP:journals/corr/abs-1301-0142}.

\subsection{UCI datasets}

We consider here the wine quality, the airfoil self-noise and the Parkinsons telemonitoring dataset \cite{Lichman:2013}. We partitioned  each dataset into source/target domain as follows:

	\begin{itemize}
    
    \item \textbf{Wine quality:} The source domain consists of only red wines, while the target domains of only white wines.
    
        \item \textbf{Airfoil noise:} We picked the second feature and computed its median over the whole dataset. The data points that have their second feature value smaller than that median will belong to the source and the rest belongs to the target domain. 

\item 	\textbf{Parkinson telemonitoring: }	
The source domain consists of male patients only, while the target domain contains female patients only. The task here is to predict the total UPDRS scores for female records.

	\end{itemize}
 To optimize the hyper-parameters $p, \lambda$, we randomly selected 20\% of the training source samples and used them as the validation set. We chose the optimal $p_*, \lambda_*$ that minimize the normalized mean squared error (NMSE) over the validation set. We then evaluated the learned models over the target testing data.  Tab.~\ref{tb:uci} shows  the average NMSE over 10 trials of all models. The last row of Tab.~\ref{tb:uci} indicates the average rank of each model over three datasets. Here the best model has the rank of 1, the second best the rank of 2 and so on. We report the average rank over the 2 datasets. Based on the ranks obtained, it is clear that the proposed (adapted) models - DA-GPR and DA-CR - can significantly improve the base regressors GPR and GCR, respectively, while outperforming the related feature adaptation approaches.\footnote{We do not report the CORAL-GP as it was consistently outperformed by CORAL-GCR on all tasks.}

\subsection{Human emotion estimation}
    
In the next experiment, we evaluate the proposed methods on the more challenging real-world data for the task of human emotion estimation (continuous valence and arousal dimensions) from the  audio-visual recordings. Specifically, we employ the 2016 AVEC RECOLA database (db) \cite{DBLP:conf/mm/ValstarGSRLTSSC16}. This db consists of 27 people  performing different collaboration tasks in a video conference. The videos are annotated by 6 annotators in terms of valence and arousal levels (continuous over time), used to obtain the gold standard, and which we consider as output in our model. We employ the audio and video features provided by the db creators, as these features have been shown to be the good predictors of target outputs \cite{DBLP:conf/mm/ValstarGSRLTSSC16}. As this data is part of the challenge, no ground-truth labels for the test partition are given. Thus, in our experiments, we use the data of 18 subjects from the training and development partitions as follows. We perform 3-fold person-independent cross-validation (6 train, 6 validation and 6 test subjects). As the evaluation measure, we use the concordance correlation coefficient (CCC) between the predicted output $\hat{y}$ and the ground-truth $y$, which is given by \cite{DBLP:conf/mm/ValstarGSRLTSSC16}:

	\begin{equation}CCC(\hat{y}; y) = \frac{2 cov(\hat{y} ; y) }{ var(\hat{y}) +var(y)) + (E(\hat{y})-E(y))^2}\end{equation}

Because we use non-parametric regression models, to speed up the computations, we randomly subsample 2000 training data from the source. We used these training data to learn the optimal feature transformations in the proposed DA, as well as CORAL and CT approaches. For our approach, we also used 2000 data from the validation set to optimize the model hyper-parameters, i.e $p$ and $\lambda$ while learning the feature transformation. The optimal hyper-parameters are then employed to evaluate, e.g., DA-GCR, over randomly selected 2000 testing points (again to avoid computational complexity of the models). However, to avoid the bias in the data used for evaluation, this procedure was repeated 10 times and the average CCC values with their standard deviation are reported. We report the performance of all models based on their predictive abilities for arousal/valence in Tab.\ref{tb:avec_arousal}-\ref{tb:avec_valence}. The last row of each table is the average rank of each model. Based on those rankings, our proposed models DA-GCR and DA-GPR outperformed their counterparts and the compared feature adaptation schemes. This evidences that our novel feature mapping not only helps to reduce the dissimilarity between two domains but also retains the most discriminative information for the regression task. 
    
The compared domain adaptation models, CORAL and CT, in most cases did not improve GCR and GPR, but even downgraded them. We attribute this to the fact that CORAL considers only the second-order matching of distributions but assuming that the input features have Gaussian distribution, which is violated in most real-world datasets. By contrast, the proposed DA based on copulas makes this assumption only in the copula subspace (and not in the input space), where the input features are transformed so as to have Gaussian distribution. Therefore, minimizing the distribution mismatch using the second order moments via Stein distance is sufficient, resulting in more robust feature subspace for the regression task. On the other hand, in contrast to our copula-based DA approach, the standard copula transformation (CT) does not exploit the source labels during the feature learning, thus, failing to learn discriminative features. As evidenced by our results, this turns out to be of significant importance for the performance by the target regression models. Likewise, the UNPRV also fails to leverage the labels from the source domain during the feature adaptation, resulting in suboptimal features for the  regression models. Finally, by comparing the performance obtained by DA-GP and DA-GCR, we note that the latter performs better in most cases. This is in part due to the fact that the copula-based regression models not only the relationships between the input features and the output, but also accounts for feature dependencies via the copula functions - resulting, in these adaptation tasks, in more robust regression model than GPs.

    \begin{table*}[tbh]
	\begin{center}
    \caption{Comparison among models for prediction of \textbf{arousal} using different modality features:  audio (Audio), video appearance (V-A), and video geometry (V-G).  \textcolor{red}{Red} indicates the best result, followed by the second best in \textcolor{blue}{blue} based on CCC.}
			\label{tb:avec_arousal}
			{\footnotesize
				\hfill{}
				\begin{tabular}{lccccccccc} 
					\toprule
					\textbf{Arousal} & GPR & GCR & CT-GPR & CT-GCR  & NPRV & UNPRV & CORAL-GCR  & DA-GPR  & DA-GCR \\
					\midrule
					\textbf{Audio} & 0.496(0.01) & \textcolor{blue}{ \textbf{0.644(0.01)}} & 0.573(0.02) &0.545(0.02) & 0.595 (0.01) & 0.595(0.01)   & 0.532(0.02)& 0.625(0.02) & \textcolor{red}{\textbf{0.670(0.01)}}   \\ 
					
					\textbf{V-A} &\textcolor{red}{\textbf{0.388(0.01)}} & 0.306(0.02) & 0.336(0.02) & 0.301(0.02) & 0.135(0.01) & 0.135(0.01) & 0.316(0.02) & \textcolor{blue}{\textbf{0.387(0.01)}} & 0.368(0.03) \\  
                    
					\textbf{V-G} &  \textcolor{red}{ \textbf{0.322(0.01)}} & 0.227(0.02) & \textcolor{blue}{\textbf{0.241(0.02)}} & 0.175(0.02) & 0.131(0.02) & 0.020(0.002)  & 0.231(0.02) & 0.287(0.03) & 0.239(0.02)  \\  
                    \midrule
                    
                    \textbf{Av.Rank} & 3.7 & 4.7  & 4.3  & 7.0 & 7.0  & 7.3 & 6.0 & \textcolor{red}{\textbf{2.3}}  & \textcolor{blue}{\textbf{2.7}}     \\                                  \bottomrule
			\end{tabular}}
			\hfill{}
	\end{center}
\end{table*}

\begin{table*}[tbh]
	\begin{center}
    \caption{Comparison among models for prediction of \textbf{valence} using different modality features:  audio (Audio), video appearance (V-A), and video geometry (V-G).  \textcolor{red}{Red} indicates the best result, followed by the second best in \textcolor{blue}{blue} based on CCC.}
			\label{tb:avec_valence}
			{\footnotesize
				\hfill{}
				\begin{tabular}{lccccccccc} 
					\toprule
					\textbf{Valence} & GPR & GCR & CT-GPR & CT-GCR  & NPRV & UNPRV & CORAL-GCR  & DA-GPR  & DA-GCR \\
					\midrule
					\textbf{Audio} & 0.315 (0.02) & \textcolor{blue}{\textbf{0.366(0.02)}} &  0.262(0.02)  & 0.302(0.03) & 0.25(0.01 ) & 0.25(0.01) & 0.365(0.03) &0.206 (0.02)  & \textcolor{red}{\textbf{0.373 (0.03)}}  \\ 
        \textbf{V-A} & 0.412 (0.01) & 0.353(0.02)   & \textcolor{blue}{ \textbf{0.437(0.01)} } & 0.356( 0.02) & 0.120 (0.02) &0.124(0.02) & 0.378(0.02) &  \textcolor{red}{\textbf{0.478(0.02)}} &0.421 (0.02)\\             
\textbf{V-G} & 0.337 (0.01)& 0.465(0.02) & 0.173 (0.01) & 0.348(0.02) & 0.083(0.01) & 0.009 (0.01) & 0.326 (0.1) &  \textcolor{red}{\textbf{0.600 (0.01) }}& \textcolor{blue}{\textbf{0.566 (0.01)}}  \\	
   \midrule
 \textbf{Av.Rank} & 4.3 & 4.0  & 5.0  & 5.0 & 8.5  & 7.8 & 4.7 & \textcolor{blue}{ \textbf{3.7}}  & \textcolor{red}{\textbf{2.0}}     \\

\bottomrule
			\end{tabular}}
			\hfill{}
	\end{center}
\end{table*}

    \section{CONCLUSIONS }
	\label{sec:copyright}
	
	We proposed a novel domain adaptation framework based on copula functions. Compared to existing feature transformation methods for domain adaptation, the two main advantages of our approach are that it does not assume Gaussian distribution in the input features, and that it exploits the source labels during the adaptation of the source to target features. As we showed in  experiments on several benchmark datasets for domain adaptation and a real-world dataset for human emotion estimation,  our adaptation approach can lead to significant performance improvements over existing methods.
	
\section{Acknowledgments}
The work of O. Rudovic is funded by the European Union H2020, Marie Curie Action - Individual Fellowship no. 701236 (EngageMe). The work of V. Pavlovic is funded by the National Science Foundation under Grant no. IIS1555408.


	\small
	\bibliography{refs}
	\bibliographystyle{IEEEbib}

\end{document}